%% file: cvpr_p.tex
\documentclass[10pt,twocolumn,letterpaper]{article}

\usepackage{cvpr}
\usepackage{time}
\usepackage[utf8]{inputenc} 
\usepackage[T1]{fontenc}    
\usepackage{url}            
\usepackage{booktabs}       
\usepackage{amsfonts}       
\usepackage{nicefrac}       
\usepackage{microtype}      
\usepackage{multirow}       
\usepackage{enumitem}
\setlist[2]{noitemsep}

\usepackage{graphicx} 
\usepackage{amsmath}
\usepackage{amssymb}
\usepackage{subcaption}
\usepackage{enumitem}
\usepackage[sort,nocompress]{cite}
\usepackage[pagebackref=true,breaklinks=true,letterpaper=true,colorlinks,citecolor=blue,bookmarks=false]{hyperref}

\usepackage{algorithm}
\usepackage[noend]{algpseudocode}

\usepackage{color}

\input{macros}

\cvprfinalcopy 

\def\cvprPaperID{1339} 

\ifcvprfinal\fi

\title{Fast Video Classification via Adaptive Cascading of Deep Models}
\author{Haichen Shen, Seungyeop Han$^*$, Matthai Philipose$^\dagger$, Arvind Krishnamurthy\\
  {\small University of Washington, Rubrik, Inc.$^*$, Microsoft Research$^\dagger$}\\
  {\tt\small \{haichen, arvind\}@cs.washington.edu, seungyeop.han@rubrik.com$^*$, matthaip@microsoft.com$^\dagger$}
}

\begin{document}

\maketitle
\thispagestyle{empty}

\input{cvpr/abstract}

\input{cvpr/intro}

\input{cvpr/relwk}
\input{cvpr/video}
\input{cvpr/cnn}
\input{cvpr/decision}
\input{cvpr/eval}
\input{cvpr/conclusion}

\section*{Acknowledgements}
\noindent
We thank anonymous reviewers for their helpful comments.
This work was funded partially by Google, Huawei, and NSF (grant: CNS-1614717).


{\small
\bibliographystyle{ieee}
\bibliography{cvpr_p}
}

\newpage
\appendix
\input{cvpr/appendix}

\end{document}

%% file: macros.tex
\definecolor{Orange}{rgb}{1,0.5,0}
\definecolor{Gray}{rgb}{0.5,0.5,0.5}
\definecolor{LtGray}{rgb}{0.66,0.66,0.66}
\definecolor{Red}{rgb}{0.8,0,0}
\definecolor{Blue}{rgb}{0,0,0.9}
\definecolor{Green}{rgb}{0,0.33,0}

\newcommand{\remove}[1]{}

\def\Snospace~{Section~{}}

\newcommand{\squishlist}{
   \begin{list}{$\bullet$}
    { \setlength{\itemsep}{0pt}      \setlength{\parsep}{0pt}
      \setlength{\topsep}{3pt}       \setlength{\partopsep}{0pt}
      \setlength{\leftmargin}{1em} \setlength{\labelwidth}{1em}
      \setlength{\labelsep}{0.5em} } }

\newcommand{\squishnone}{
   \begin{list}{}
    { \setlength{\itemsep}{0pt}      \setlength{\parsep}{0pt}
      \setlength{\topsep}{3pt}       \setlength{\partopsep}{0pt}
      \setlength{\leftmargin}{0em} \setlength{\labelwidth}{1em}
      \setlength{\labelsep}{0.5em} } }

\newcommand{\squishlistcirc}{
   \begin{list}{$\circ$}
    { \setlength{\itemsep}{0pt}      \setlength{\parsep}{0pt}
      \setlength{\topsep}{3pt}       \setlength{\partopsep}{0pt}
      \setlength{\leftmargin}{1.5em} \setlength{\labelwidth}{1em}
      \setlength{\labelsep}{0.5em} } }

\newcommand{\squishlisttwo}{
   \begin{list}{$\bullet$}
    { \setlength{\itemsep}{0pt}    \setlength{\parsep}{0pt}
      \setlength{\topsep}{0pt}     \setlength{\partopsep}{0pt}
      \setlength{\leftmargin}{2em} \setlength{\labelwidth}{1.5em}
      \setlength{\labelsep}{0.5em} } }

\newcommand{\squishend}{
    \end{list} \vspace{5pt} }



%% file: cvpr/abstract.tex
\begin{abstract}
\noindent
Recent advances have enabled ``oracle'' classifiers that can classify
across many classes and input distributions with high accuracy
without retraining. However, these classifiers are relatively
heavyweight, so that applying them to classify video is costly. We
show that day-to-day video exhibits highly
skewed class distributions over the short term, and that these
distributions can be classified by much simpler models. We formulate
the problem of detecting the short-term skews online and exploiting models based
on it as a new sequential decision making problem dubbed the Online
Bandit Problem, and present a new algorithm to solve it. When applied to
recognizing faces in TV shows and movies, we realize end-to-end
classification speedups of  2.4-7.8$\times$/2.6-11.2$\times$ (on
GPU/CPU) relative to a state-of-the-art convolutional neural network,
at competitive accuracy.
\end{abstract}


%% file: cvpr/intro.tex

\section{Introduction}
\label{s:intro}
\noindent
Consider recognizing entities such as objects, people, scenes and
activities in every frame of video footage of day-to-day life. Such
footage may come, for instance, from the media, wearable cameras,
movies, or surveillance cameras. In principle, these entities could be
drawn from thousands of classes: many of us encounter hundreds to
thousands of distinct people, objects, scenes and activities through
our life. Over short intervals such as minutes, however, we tend to
encounter a very small subset of classes of entities. For instance, a
wearable camera may see the same set of objects from our desk at work
for an hour, a movie may focus only on cooking-related activities
through a five-minute kitchen sequence, and media footage of an event
may focus on only those celebrities participating in the event. In
this paper, we characterize and exploit such {\em short-term class
  skew} to significantly reduce the latency of classifying video using
Convolutional Neural Networks (CNNs).

Since the seminal work of Viola and Jones \cite{viola-jones} on face
detection, one of the best-known techniques to speed up
classification has been to structure the classifier as a cascade (or
tree \cite{JMLR:v15:xu14a}) of simple classifiers such that ``easy''
examples lead to early exits and are therefore classified
faster. Cascaded classifiers require that training and test data are
strongly (and identically) biased toward a small number of easy to
detect classes. In the (binary) face detection task, for example, the class
``not a face'' is both (i) by far more common than ``face'', and (ii)
often quite easy to classify via a small number of comparisons of
inexpensive Haar-style features. In fact, traditional cascades are
most applicable in {\em detection} tasks \cite{DollarPAMI14pyramids},
where the background is both much more common and easier to classify than
the foreground.

Distinct from the (two-class) detection setting in traditional cascading,
recent advances in convolutional neural networks (CNNs)
\cite{google-lenet, mit-scene-nips, vgg-face} have opened up the
possibility of using a
single, pre-trained ``oracle'' classifier to {\em recognize} thousands of
classes such as people, objects and scenes. When training such oracle
classifiers, such as GoogLeNet \cite{google-lenet} of VGGFace
\cite{vgg-face}), a small number of classes
do not usually dominate the training set: for broad applicability, the
classifier is trained assuming that all classes are more or less
equally likely. Even if such a skew toward such small classes existed,
there is no {\em a priori} reason that these dominant classes are fast
to classify. It may seem therefore that
cascading is not a promising optimization for improving the speed of
entity recognition in video via CNNs.


We demonstrate, however, that for many recognition tasks, {\em
  day-to-day video often exhibits
significant  short-term skews} in class distribution. We present
measurements on a diverse set of videos that show, for instance, that
in over 90\% of 1-minute windows, at least 90\% of objects interacted
with by humans belong to a set of 25 or fewer objects. The underlying
ImageNet-based recognizer, on the other hand, can recognize up to 1000
objects. We show that similar skews hold for faces and scenes in
videos.

Even if such skew exists, to our knowledge, it
has not been shown that distributions skewed toward small sets of
classes can be classified accurately by simpler CNNs than uniformly
distributed ones. We therefore also
demonstrate that {\em when class distribution is highly skewed,
  ``specialized'' CNNs trained to classify inputs from this
  distribution can be much more compact} than
the oracle classifier. For instance, we present a CNN that executes
200$\times$ fewer FLOPs than the state-of-the art VGGFace \cite{vgg-face}
model, but has comparable accuracy when over
50\% of faces come from the same 10 or fewer people.  We
present similar order-of-magnitude faster specialized CNNs for
 object and scene recognition.


Given the ability to produce fast, accurate versions of CNNs
specialized for particular test-time skews, we seek to estimate
the (possibly non-stationary) skew at test-time, produce a
specialized model if appropriate, exploit the model as long as the
skew lasts, detect when the skew disappears and then revert to the oracle
model. As with standard ``bandit''-style sequential decision-making
problems, the challenge is in balancing  exploration (i.e.,
using the expensive oracle to estimate the skew) with  exploitation
(i.e., using a model specialized to the current best available
estimate of the skew). We formalize this problem as the Oracle Bandit
Problem and propose a new exploration/exploitation-based algorithm we dub
Windowed $\epsilon$-Greedy (WEG) to address it.

Using a combination of synthetic data and real-world videos, we
empirically validate the WEG algorithm. In particular, we show that WEG
can reduce the end-to-end classification overhead of face recognition on
TV episodes and movies by 2.4-7.8$\times$ relative to unspecialized
classification using the VGGFace classifier on a GPU (2.6-11.2$\times$
on a CPU). We show via synthetic data that similar gains are to be had
on object and scene recognition as well. We provide a detailed
analysis of WEG's functioning, including an accounting of how much
its key features contribute. To  our knowledge our
system is the first to use test-time sequential class skews in video to produce
faster classifiers.

%% file: cvpr/relwk.tex
\section{Related work}
\label{s:relwk}
\noindent
There is a long line of work on cost-sensitive classification, the
epitome of which is perhaps the cascaded classification work of Viola
and Jones \cite{viola-jones}. The essence of this line of work
\cite{yangTestCostTKDE06, xuCostSensitiveICML13} is to
treat classification as a sequential process that may exit early if it
is confident in its inference, typically by learning sequences
that have low cost in expectation over training
data. Recent work \cite{LiLSBH15} has even proposed cascading CNNs as
we do. All these techniques assume 
that testing data is  i.i.d.\  (i.e., not sequential), that all
training happens before any testing, and rely on skews in {\em training}
data to capture cost structure. As such, they are not equipped to
exploit short-term class skews in test data.

Traditional sequential models such as probabilistic models
\cite{wuICCV07, desaiECCV12, pirsiavashCVPR12} and
Recurrent Neural Networks (RNNs) \cite{lrcn2014, KarpathyCVPR14} are aimed at classifying instances
that are not independent of each other. Given labeled sequences as
training data, these techniques learn more accurate classifiers than
those that treat sequence elements as independent.  However, to 
 our knowledge, none of these approaches produces classifiers
that yield {\em less expensive} classification in response to
favorable inputs, as we do.

Similar to adaptive cascading, online learning methods
\cite{Valko10, harICCV11, kveton13} customize
models at test time. For training, they use labeled data from a
sequential stream that typically contains both labeled and unlabeled
data. As with adaptive cascading, the test-time cost of incrementally
training the model in these systems needs to be low.  A fundamental
difference in our work is that we make no assumption that our input
stream is partly labeled. Instead, we assume the availability of a
large, resource-hungry model that we seek to ``compress'' into a
resource-light cascade stage.  

Estimating distributions in sequential data and exploiting it is the
focus of the multi-armed bandit (MAB) community
\cite{Auer02, Lai85}. The Oracle Bandit
Problem (OBP) we define differs from the classic MAB setting in that
in MAB the set of arms over which exploration and exploitation happen
are the same, whereas in OBP only the oracle ``arm'' allows
exploration whereas specialized models allow exploitation. Capturing
the connection between these arms is the heart of the OBP
formulation. Our Windowed $\epsilon$-Greedy algorithm is strongly
informed by the use of windows in \cite{garivier2008upper} to handle
non-stationarities and the well-known \cite{Sutton98} $\epsilon$-greedy  scheme
to balance exploration and exploitation.

Finally, much recent work has focused on reducing the resource
consumption of (convolutional) neural networks
\cite{DBLP:journals/corr/HanMD15, xnor-arxiv, ba2014deep, NIPS2015_5784}. These techniques are
oblivious to test-time data skew and are complementary to
specialization. We expect that even more pared-down versions of these optimized
models will provide good accuracy when specialized at test-time.



%% file: cvpr/video.tex

\section{Class skew in day-to-day video}
\label{s:video}



\noindent
Specialization depends on skew (or bias) in the temporal distribution
of classes presented to the classifier. In this section, we analyze
the skew in videos of day-to-day life culled from YouTube. We
assembled a set of 30 videos of length 3 minutes to 20 minutes from
five classes of daily activities: socializing, home repair, biking
around urban areas, cooking, and home
tours.
We expect this kind of footage to come from a variety of sources such as movies, amateur productions of the kind that dominate YouTube and wearable videos.

We sample one in three frames uniformly from these videos and apply
state-of-the-art face (derived from \cite{vgg-face}), scene
\cite{mit-scene-nips} and object recognizers \cite{vggnet} to every
sampled frame. Note that these ``oracle'' recognizers can recognize up
to 2622 faces, 205 scenes and 1000 objects respectively. For face
recognition, we record the top-scoring label for each face detected,
and for the others, we record only the top-scoring class on each
frame. For object recognition in particular, this substantially
undercounts objects in the scene; our count (and specialization)
applies to applications that identify some distinctive subset of
objects (e.g., all objects ``handled'' by a person). We seek to
compare these numbers to the number of distinct recognized faces,
scenes and objects that dominate ``epochs'' of $\tau$ = 10 seconds, 1
minute and 3 minutes.  

\begin{figure}[t]
\vspace{-.15in}
\centering
\captionsetup{font=small}
\begin{subfigure}[t]{0.23\textwidth}
\includegraphics[width=\textwidth]{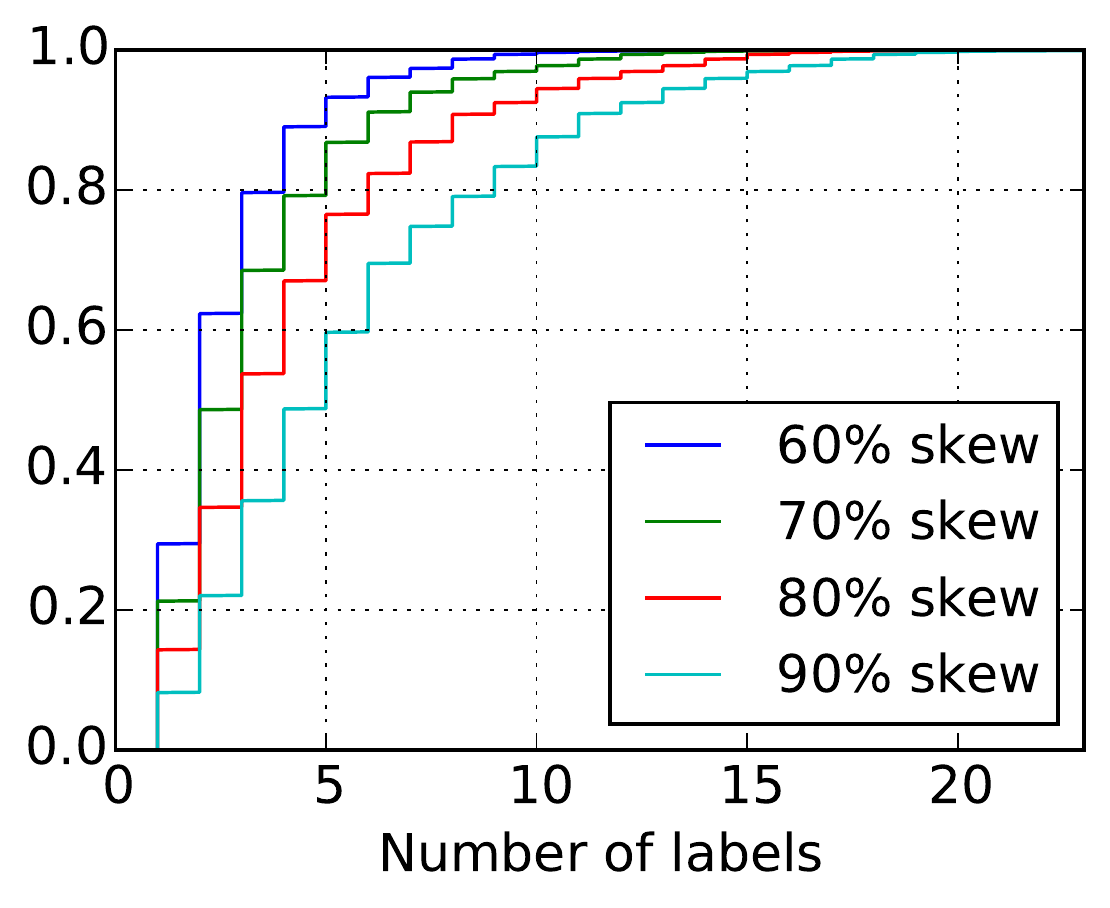}
\caption{10s segments, objects}
\end{subfigure}
\begin{subfigure}[t]{0.23\textwidth}
\includegraphics[width=\textwidth]{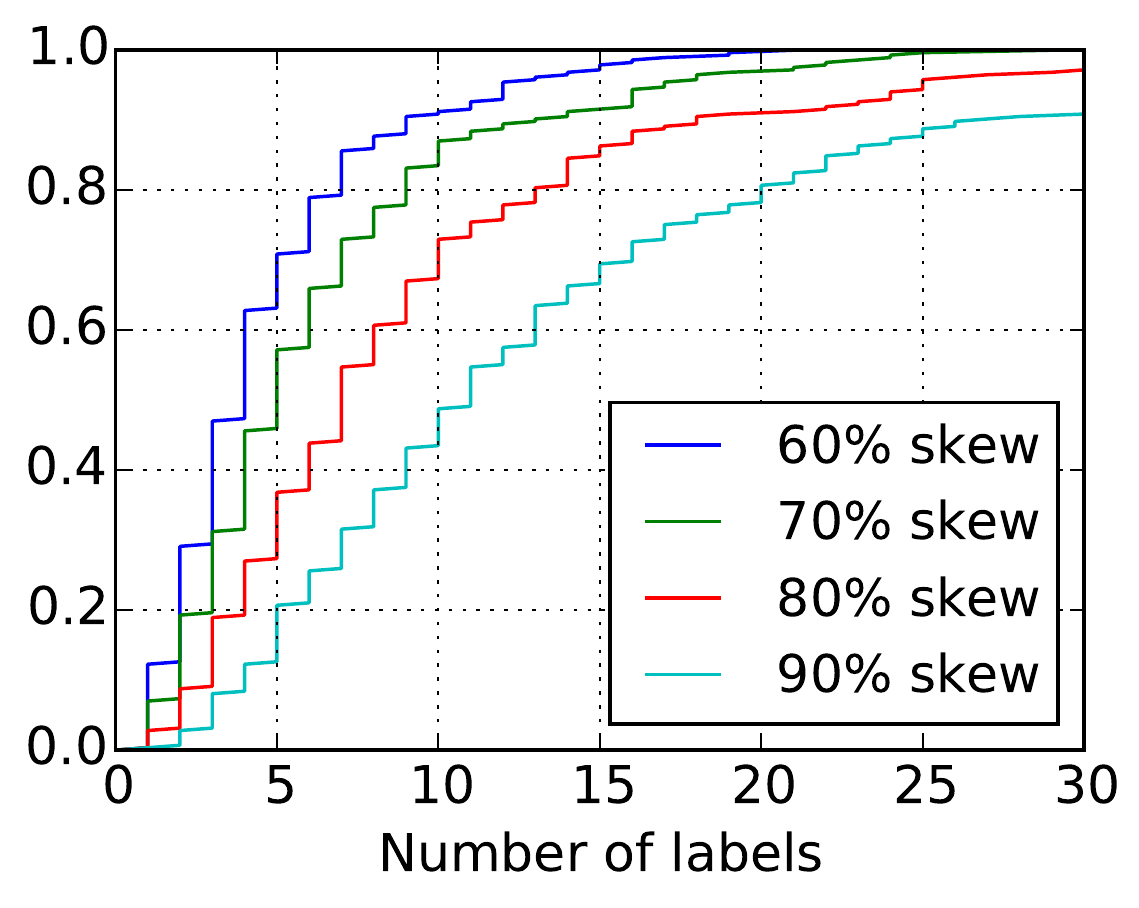}
\caption{1 min seg., objects}
\end{subfigure}
~
\begin{subfigure}[t]{0.23\textwidth}
\includegraphics[width=\textwidth]{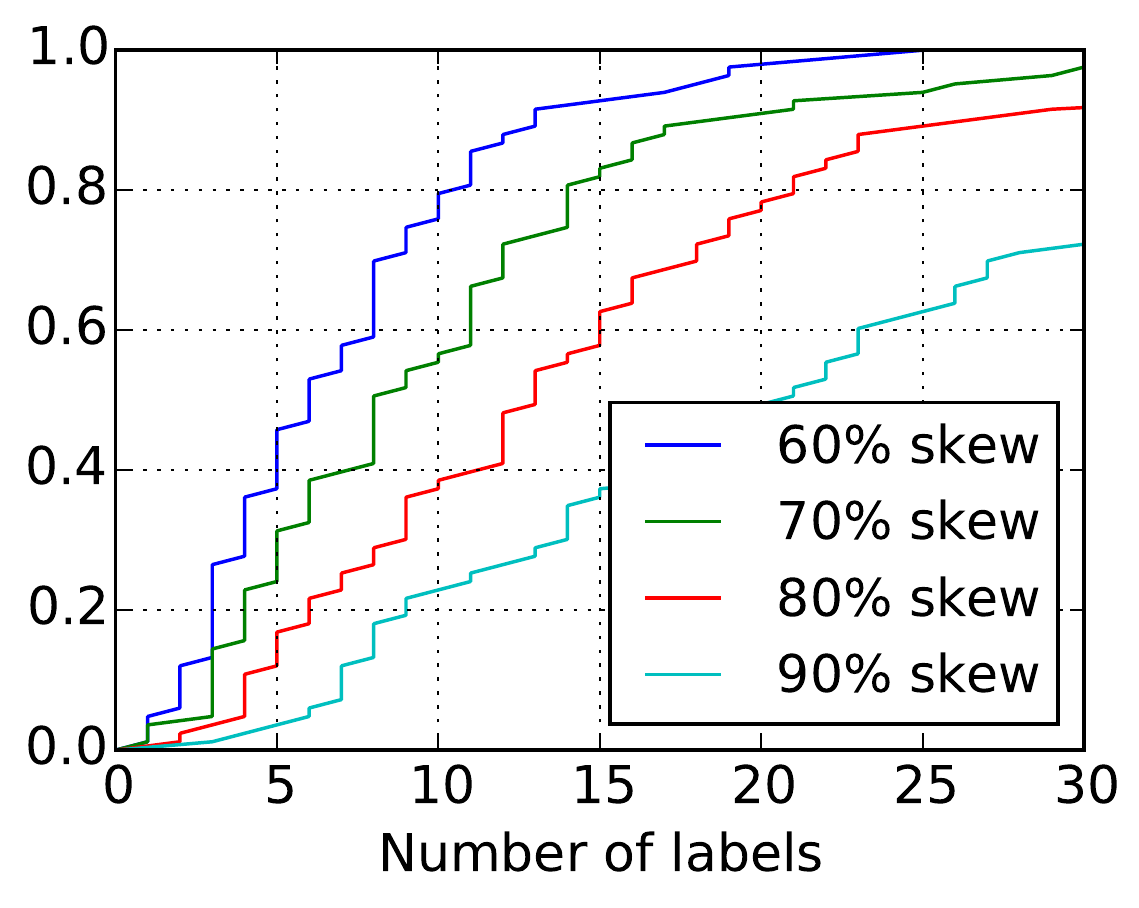}
\caption{3 min seg., objects}
\end{subfigure}
\begin{subfigure}[t]{0.23\textwidth}
\includegraphics[width=\textwidth]{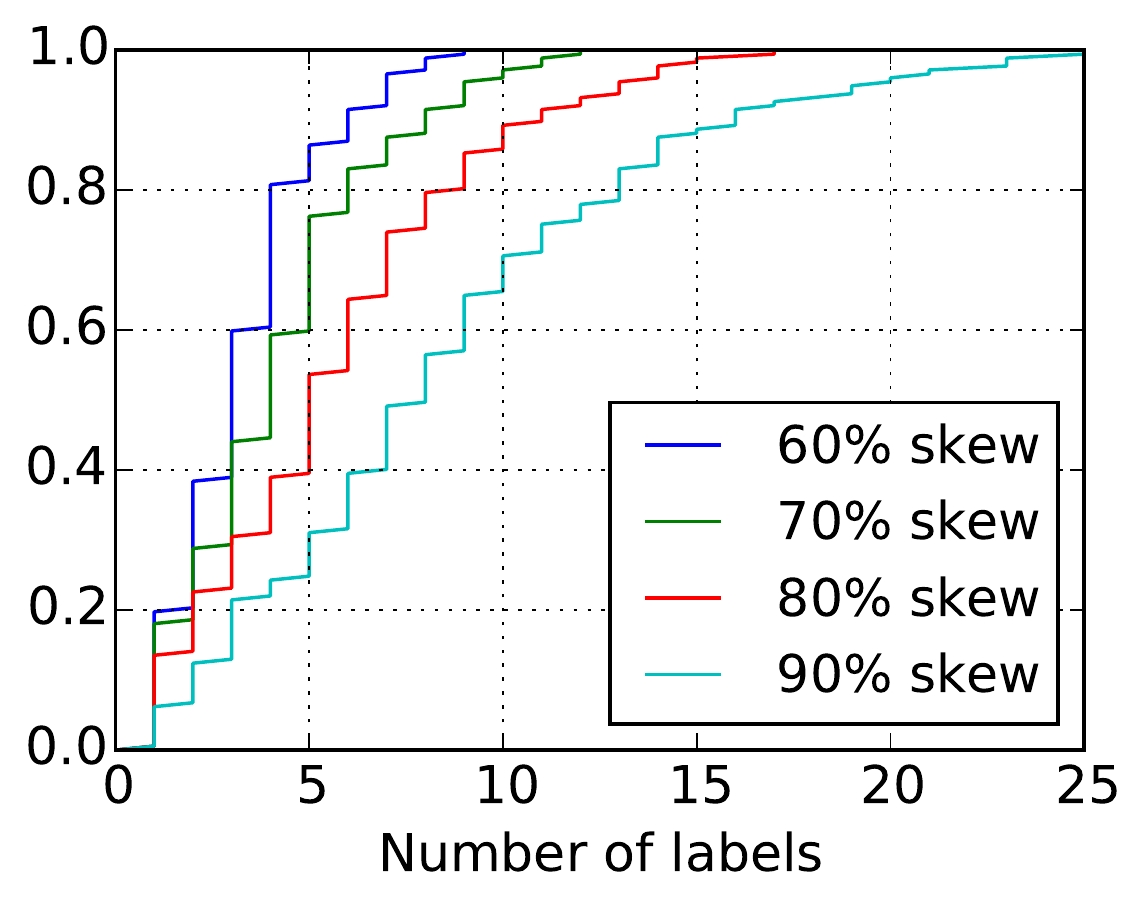}
\caption{1 min seg., scenes}
\end{subfigure}
\vspace{-.1in}
\caption{Temporal skew of classes in day-to-day video.}
\label{fig:video_scene_skew}
\vspace{-15pt}
\end{figure}

\autoref{fig:video_scene_skew} shows the results for object
recognition and scene recognition. We partition the sequence of frames
into segments of length $\tau$ and show one plot per segment length. 
Each line in the plot corresponds to percentage skew $s \in \{60, 70,
80, 90\}$. Each line in the plots shows the cumulative distribution
representing the fraction of all segments where $n$ labels comprised
more than $s$ percent of all labels in the segment. For instance, for
10-second segments (\autoref{fig:video_scene_skew}(a)), typically
roughly 100 frames, 5 objects comprised 90\% of all objects in a
segment 60\% of the time (cyan line), whereas they comprise 60\% of
objects 90\% of the time (dark blue).

In practice, detecting skews and training models to exploit them
within 10 seconds is often challenging. As figures (b) and (c) show,
the skew is less pronounced albeit still very significant for longer
segments. For instance, in 90\% of 3-minute segments, the top 15
objects comprise 90\% of objects seen. The trend is similar with faces
and scenes, with the skew significantly more pronounced, as is
apparent from comparing figures (b) and (d); e.g. the cyan line in (d)
dominates that in (b). We expect that if we ran a hand-detector and
only recognized objects in the hand (analogously to
recognizing detected faces), the skew would be much sharper. 

Specialized models must exploit skews such as these to deliver
appreciable speedups over the oracle. Typically, they should be
generated in much less than a minute, handle varying amounts of skew
gracefully, and deliver substantial speedups when inputs belong to
subsets of 20 classes or fewer out of a possible several hundred in
the oracle.



%% file: cvpr/cnn.tex
\section{Specializing Models}
\label{s:cnn}




\begin{table}[t]
\centering
\scriptsize
\setlength{\tabcolsep}{3pt}
\input{cvpr/table_cnn}
\caption{Oracle classifiers versus compact classifiers in top-1
  accuracy, number of FLOPs, and execution time.
  Execution time is
  feedforward time of a single image without batching on Caffe
  \cite{jia2014caffe},  a Linux server with a 24-core
   Intel Xeon E5-2620 and an NVIDIA K20c GPU.}
\label{tab:cnns}
\end{table}

\noindent
In order to exploit skews in the input, we cascade the expensive but
comprehensive {\em oracle model} with a (hopefully much) less expensive
``compact'' model. This {\em cascaded classifier} is designed so that if its
input belongs to the frequent classes in the incoming distribution it will
return early with the classification result of compact model, else it will
invoke the oracle model. Thus if the skew dictates that $n$ frequent classes, or
{\em dominant classes}, comprise percentage $p$ of the input, or {\em skew},
model execution will cost the overhead of just executing compact model roughly
$p$\% of the time, and the overhead of executing compact model and oracle
sequentially the rest of the time. When $p$ is large, the lower cost compact
model will be incurred with high probability.

To be more concrete, we use state of the art convolutional neural networks
(CNNs) for oracles. In particular, we use the GoogLeNet~\cite{google-lenet} as
our oracle model, for object recognition; the VGG Net 16-layer version for scene
recognition~\cite{mit-scene-nips}; and the VGGFace network~\cite{vgg-face} for
face recognition. The compact models are also CNNs. For these, we use
architectures derived from the corresponding oracles by systematically (but
manually) removing layers, decreasing kernel sizes, increasing kernel strides,
and reducing the size of fully-connected layers. The end results are
architectures (O[1|2] for objects, S[1|2] for scenes and F[1|2] for faces) that
use noticeably less resources (\autoref{tab:cnns}), but also yield significantly
lower average accuracy when trained and validated on unskewed data, i.e., the
same training and validation sets for oracle models. For instance, O1 requires roughly
4$\times$ fewer FLOPs to execute than VGGFace, but achieves roughly 70\% of its
accuracy.

\begin{figure}[t]
  \centering
  \begin{subfigure}[t]{.8\linewidth}
    \hspace{-15pt}(a) \raisebox{\dimexpr-\height+\baselineskip}{\includegraphics[width=\textwidth]{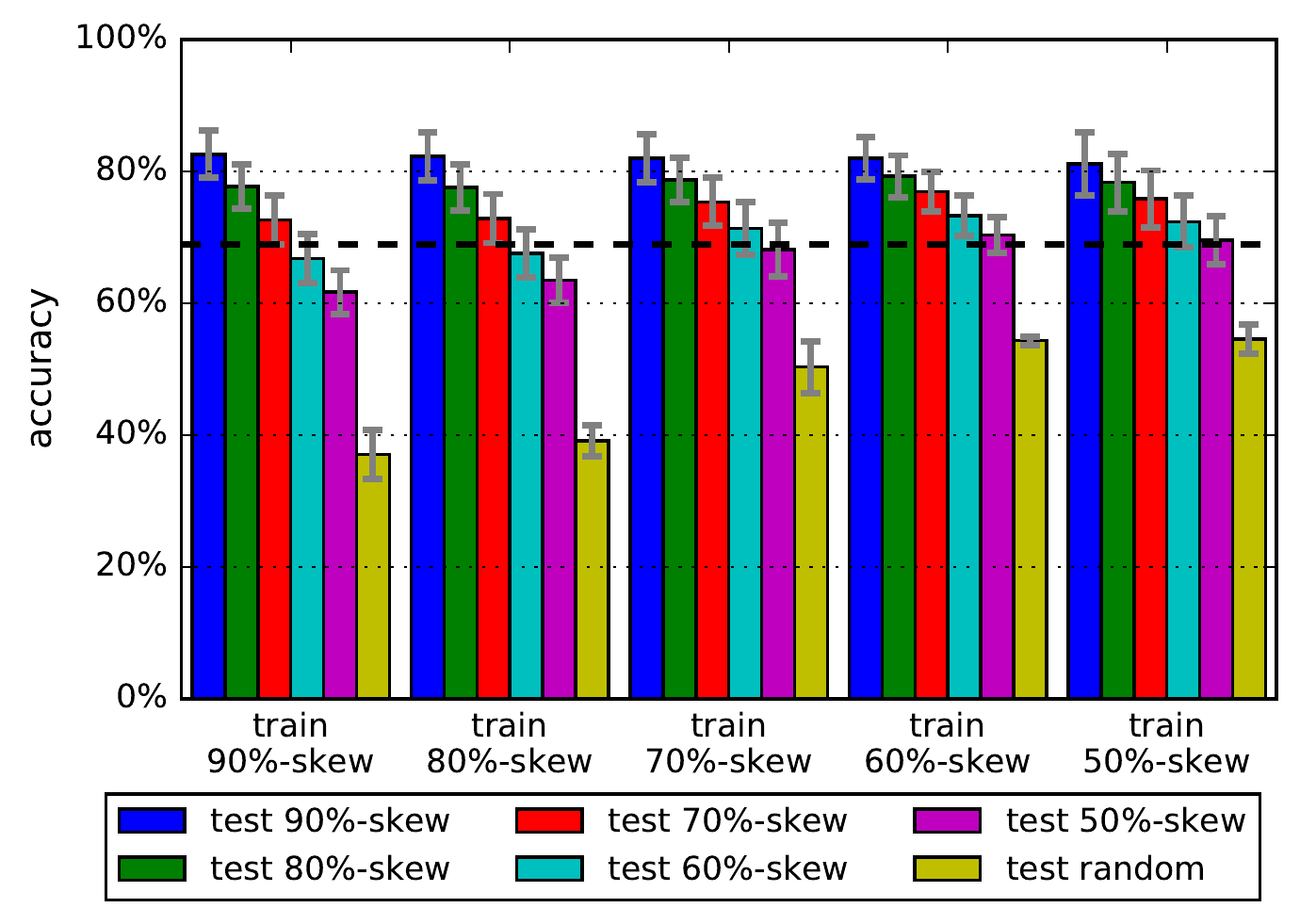}}
    \label{fig:o3-skew-acc}
  \end{subfigure}
  ~
  \begin{subfigure}[t]{.8\linewidth}
    \hspace{-15pt}(b) \raisebox{\dimexpr-\height+\baselineskip}{\includegraphics[width=\textwidth]{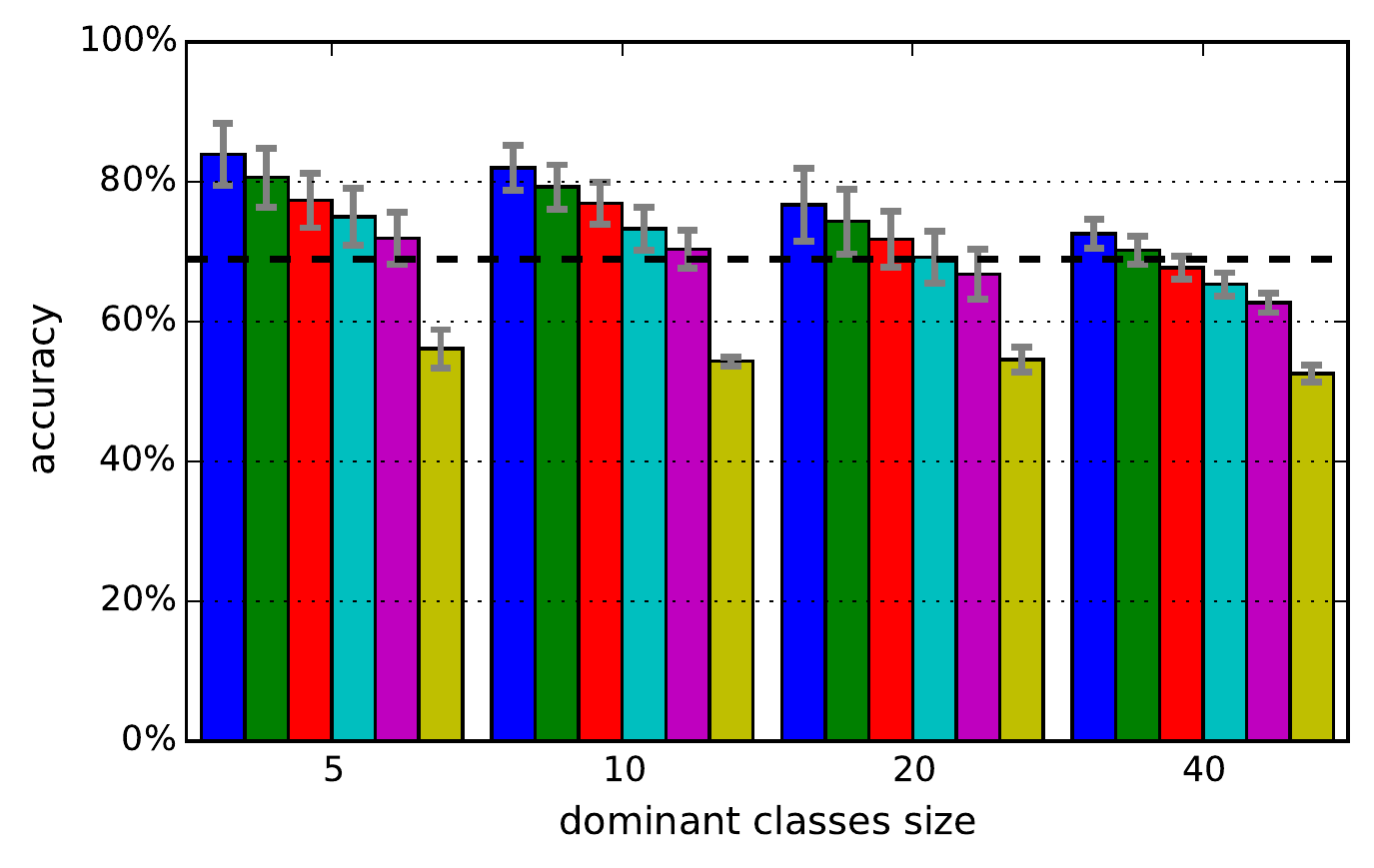}}
    \label{fig:o3-vary-ncls}
  \end{subfigure}
  \caption{(a) When compact model O1 is trained by various skews and cascaded
    with the oracle, accuracy of the cascaded classifier tested by various skews
    for 10 dominant classes; (b) accuracy of O1 trained by 60\% skew and tested
    by various skews for different number of dominant classes. The dashed line
    shows the accuracy of GoogLeNet as a baseline. All experiments repeated 5x
    with randomly selected dominant sets.}
  \label{fig:spec}
  \vspace{-15pt}
\end{figure}

\begin{figure}[t]
  \centering
  \includegraphics[width=.8\linewidth]{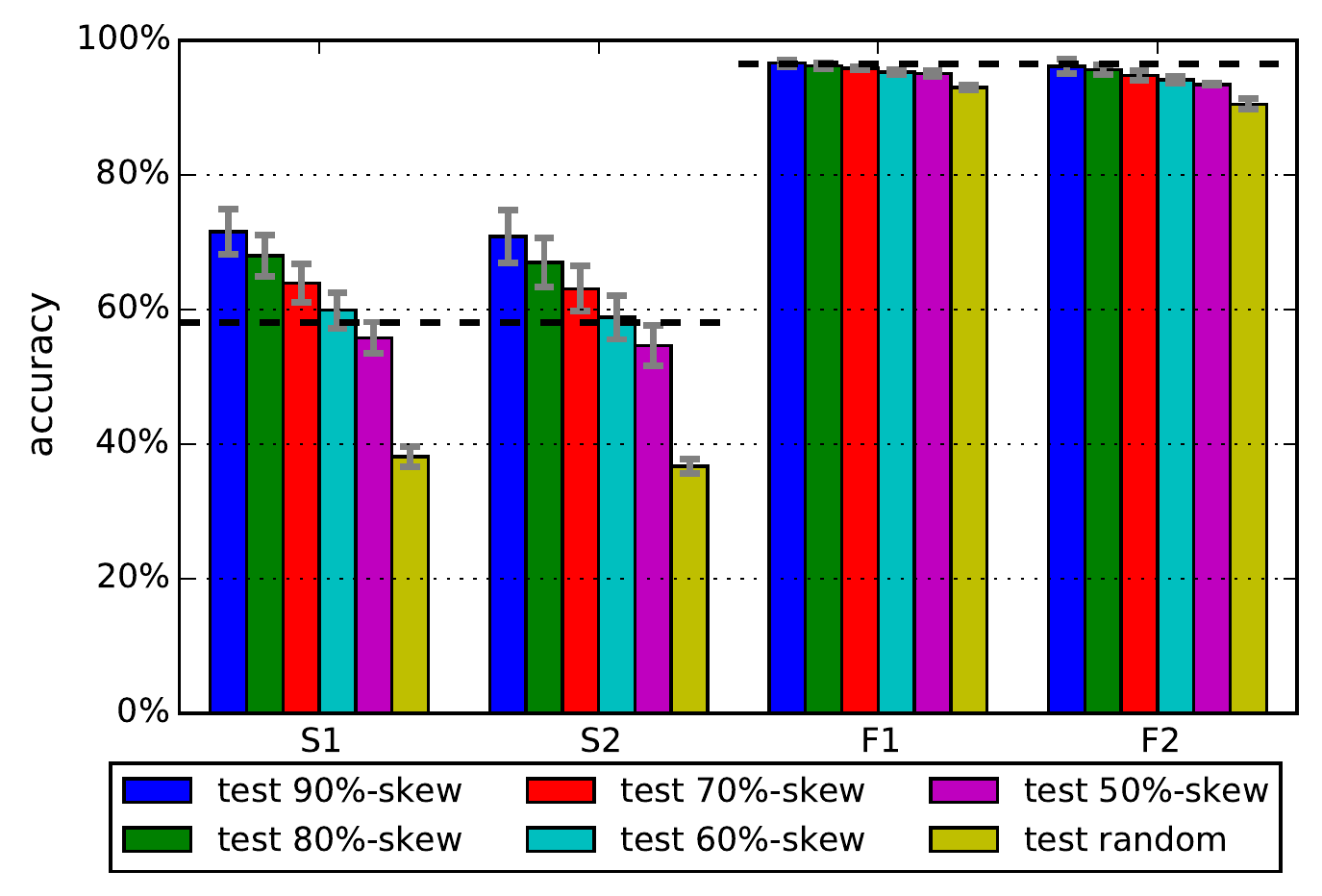}
  \caption{Accuracy of scene classifiers trained by 70\% fixed skew and 10
    dominant classes and face classifiers trained by 50\% fixed skew and 10
    dominant classes. Dashed lines show the accuracy of the oracle classifier
    for scene and face recognition task.}
  \label{fig:spec-others}
  \vspace{-15pt}
\end{figure}

However, in our approach, we train these compact models to classify {\em skewed}
distributions observed during execution, denoted by {\em specialized
  classifier}, and their performance on skewed distributions is the critical
measure. In particular, to generate a specialized model, we create a new
training dataset with the data from the $n$ dominant classes of the original
data, and a randomly chosen subset from the remaining classes with label
``other'' such that the dominant classes comprise $p$ percent of the new data
set. We train the compact architecture with this new dataset.

\autoref{fig:spec} shows how compact models trained on skewed data and cascaded
with their oracles perform on validation data of different skews.
\autoref{fig:spec}(a) analyzes the case where $n=10$, for various combinations
of training and validation skews for model O1. Recall from \autoref{tab:cnns}
that O1 delivers only 70\% of its accuracy on unskewed inputs. However, when
training and testing is on skewed inputs, the numbers are much more favorable.
When O1 is trained on $p$=90\% skewed data with $n$=10 dominant classes, it
delivers over 84\% accuracy on average (the left-most dark-blue bar). This is
significantly {\em higher} than the oracle's average of 68.9\% (top-1 accuracy),
denoted by the horizontal black line.
We also observed from \autoref{fig:spec}(a) that when O1 is trained
on 60\% skewed data, the cascaded classifier maintains high accuracy across
 a wide range of testing skews from 90\% to 50\%.
Therefore, in what follows, we use 60\% skew as {\em fixed training
  skew} to specialize object compact models in the rest of paper
(similarly 70\% fixed skew for scene and 50\% for face).
\autoref{fig:spec}(b) shows that, where $n$ is varied for O1, the cascaded classifier
degrades very gracefully with $n$. Finally, \autoref{fig:spec-others}, which
reports similar measurements on compact models S[1|2] and F[1|2] shows
that these trends carry over to scene and face recognition.


Finally, we note that since skews are only evident at test-time, specialized
models must be trained extremely fast (ideally a few seconds at most). We use
two techniques to accomplish this. First, before we begin processing any inputs,
we train all model architectures on the full, unskewed datasets of their
oracles. At test time, when the skew $n, p$ and the identity of dominant classes
is available, we only retrain the top (fully connected and softmax) layers of
the compact model. The lower layers, being ``feature calculation'' layers do not
need to change with skew.
Second, as a pre-processing step, we run all inputs in the training dataset
through the lower feature-calculation layers, so that when re-training the top
layers at test time, we can avoid doing so.  This combination of techniques
allows us to re-train the specialized model in roughly 4s for F1 and F2 and 14s
for O1/O2, many orders of magnitude faster than fully re-training these
models.




%% file: cvpr/table_cnn.tex
\begin{tabular}{ccccccc}
\toprule
  Task & Model & Acc.(\%) & FLOPs & CPU lat.(ms) & GPU lat.(ms) \\
\midrule
  \multirow{3}{*}{\parbox{1cm}{\centering Object\\(1000 classes)}} &
  \cite{google-lenet} & 68.9 & 3.17G & 779.3 & 11.0 \\
  & O1 & 48.9 & 0.82G & 218.2 ($\times$3.6) & 4.4 ($\times$2.5) \\
  & O2 & 47.0 & 0.43G & 109.1 ($\times$7.1) & 2.8 ($\times$3.9) \\ 
\midrule
  \multirow{3}{*}{\parbox{1cm}{\centering Scene\\(205)}} &
  \cite{mitplaces-cnn} & 58.1 & 30.9G & 2570 & 28.8 \\
  & S1 & 48.9 & 0.55G & 152.2 ($\times$16.9) & 3.36 ($\times$8.6) \\
  & S2 & 40.8 & 0.43G & 141.5 ($\times$18.2) & 2.44 ($\times$11.8) \\
\midrule
  \multirow{3}{*}{\parbox{1cm}{\centering Face\\(2622)}} &
  \cite{vgg-face} & 95.8 & 30.9G & 2576 & 28.8 \\
  & F1 & 84.8 & 0.60G & 90.1 ($\times$28.6) & 2.48 ($\times$11.6) \\
  & F2 & 80.9 & 0.13G & 40.4 ($\times$63.7) & 1.93 ($\times$14.9) \\
\bottomrule
\end{tabular}


%% file: cvpr/decision.tex
\newcommand{\ForAllIn}[2]{\ForAll{#1 \textbf{ in } #2}}

\section{Sequential Model Specialization}

\subsection{The Oracle Bandit Problem (OBP)}
\noindent
Let $x_1,x_2,\ldots,x_i,\ldots \in X = \mathbb{R}^{n}$ be a stream of
images to be classified. Let $y_1,y_2,\ldots,y_i,\ldots \in Y=
[1,\ldots,k]$ be the corresponding classification results. Let $\pi:
\mathbb{I^+} \to \mathbb{I^+}$ be a partition over the
stream. Associate the distribution $T_j$ with partition $j$, so that
each pair $(x_i,y_i)$ is sampled from $T_{\pi(i)}$. Intuitively, $\pi$
partitions, or segments, $\ldots, x_i, \ldots$ into a sequence of
``epochs'', where elements from each epoch $j$ are drawn independently
from the corresponding stationary distribution $T_j$. Thus, for the
overall series, samples are drawn from an abruptly-changing,
piece-wise stationary distribution. At test time, neither results
$y_i$ nor partitions $\pi$ are known. 

Let $h^*:X \to Y$ be a classifier, designated the ``oracle''
classifier, trained on distribution $T^*$. Intuitively $T^*$ is a
mixture of all distributions comprising the oracle's input stream:
$T^* = \sum_{j} T_j$. Let $R^*$ be the cost (e.g., number
cycles), assumed invariant
across $X$, needed to execute $h^*$ on any 
$x \in X$. At test time, on each input $x_i$, we can always consult
$h^*$ at 
cost $R^*$ to get a label $y_i$ with some (high) accuracy $a^*$.

Let $m_1,\ldots, m_M$ be {\em model architectures}, such as those
of O1, O2, S1, S2, F1 and F2 in \autoref{tab:cnns}. Suppose each
architecture $m_k$ is trained {\em offline} on $T^*$ to obtain a
``template'' classifier $h_k$.
We assume that re-targeting template
$h_k$ to a new size-$j$ set of dominant classes has a flat cost $R_0$.

Finally, for each set of dominant classes $D$, the corresponding 
{\em specialized classifier} $h_{D}$ is trained by re-targeting some
template $h_k$, using a dataset that 
draws half its examples from classes in $D$ and the
rest (with a single label ``other'') from $Y-D$. Let the cost of
executing $h_d$ be $R_{h_d}$. Chaining $h_{D}$ with $h^*$ gives a {\em
  cascaded} classifier 
$\hat{h}_{D}(x)\triangleq$ if $y = h_{D}(x) \in D$, return
$y$, otherwise return $h^*(x)$.  Note that executing $\hat{h}_{D}$
will either cost $R_{h_d}$ (in the case that the condition is true),
or  $R_{h_d} + R^*$ in the case that it is false. Given that $R_{h_d} \ll R^*$,
developing and using specialized classifiers $h_D$ can thus reduce costs
significantly. Since $x_i$ is drawn from some
distribution $T$, each classifier $\hat{h}_{D}$ also has {\em  cost}
that belongs to a corresponding distribution, which we write as
$R_{T\hat{h}_{D}}(x_i)$.



Now consider a policy (or algorithm) $P$ that, for each incoming image $x_i$
belonging to stationary distribution $T_{\pi(i)}$ as  above, selects
a classifier $\hat{h}^{(i)}_{D}$ (for some set choice of $D$), and
applies it to $x_i$. The classifier selected could also include the
oracle. The expected total cost of this policy, $R_p =
|\cup_i\{\hat{h}^{(i)}_D\}|R_0 + \Sigma_i
E_{x_i \sim T_{\pi(i)}}(R_{T_{\pi(i)}{\hat{h}^{(i)}_D}}(x_i))$. We seek a
minimal-cost policy: $P^* = \arg\min_P R_P$ that maintains average accuracy
within a threshold $\tau_a$ of oracle accuracy $a^*$.

\subsection{The Windowed $\epsilon$-Greedy (WEG) Algorithm}
\label{s:skew_detection}
\noindent
A close look at the policy cost above provides some useful intuition
on what good policies should do. First, given the high fixed cost $R_0$ of
re-targeting models as opposed to just running them, re-targeting
should be infrequent. We expect re-targeting to occur roughly once an
epoch. Second, the 
cost of running the cascade is much lower than that of running the
oracle {\em if the input $x_i$ is in the dominant class set $D$} and
higher otherwise. It is important therefore to identify promptly when a
dominant set $D$ exists, produce a specialized model $h_D$ that does
not lose too much accuracy, and revert back to the oracle model when
the underlying distribution changes and $D$ is no longer dominant. We
provide a heuristic exploration-exploitation based algorithm
(\autoref{alg:opt-alg}) based on these intuitions.

\algnewcommand{\LineComment}[1]{\(\triangleright\) #1}
\begin{algorithm}[t]
\begin{algorithmic}[1]
\scriptsize
\State $j, S_0 \leftarrow 1, []$
\Statex \LineComment{Note: $\tau_r$, $\tau_a$, $\tau_{FP}$ and $\epsilon$ below are
  hyper-parameters.}
\Statex {\bf Window Initialization Phase}
\State Repeat $w_{min}$ times \label{algo:line-sample}
\State \hspace{\algorithmicindent} $y_t \leftarrow h^*(x_t)$
\State \hspace{\algorithmicindent} $S_j \leftarrow S_j \oplus [y_t]$
\Comment{Append new sample}
\If{$||\textsc{DomClasses}(S_{j-1}), \textsc{DomClasses}(S_j)|| \leq  \tau_r$}
\Statex \Comment{dominant classes match sufficiently, old epoch continues}
\State $S_j \leftarrow S_{j-1} \oplus S_j$ \label{algo:line-merge}
\EndIf
\State $w \leftarrow |S_j|$ and go to Line \ref{algo:line-normal} \label{algo:line-end-initialization}
\Statex

\Statex {\bf Template Selection Phase}
\State $D \leftarrow \textsc{DomClasses}(\textrm{last $w$
  elements in }S_j)$ \label{algo:line-normal}
\State{Estimate acc. $a_{\hat{h}_D}$ of $\hat{h}_D$; use $p^*$
  derived from $S_j$ (\autoref{eqn:acc})} 
\If{$a_{\hat{h}_D} \geq a^* + \tau_a$} \label{algo:line-test-accuracy}
\State train specialized classifier $h_D$ on dominant classes $D$
\State go to Line \ref{algo:line-spec-init} \Comment{Exploit cascaded
  classifier $\hat{h}_D$}
\EndIf
\State $y_t \leftarrow h^*(x_t)$ \Comment{Else, continue exploring with oracle}
\State $S_j \leftarrow S_j \oplus [y_t]$
\State go to Line \ref{algo:line-normal} \label{algo:template-selection-end}
\Statex

\Statex {\bf Specialized Classification Phase}
\State $n_c, n^*, S \leftarrow 0, 0, S_j$ \label{algo:line-spec-init}  
\State $y_t, c \leftarrow \hat{h}_D(x_t)$ \Comment{exploit; $c=0|1$ if|if-not cascaded to oracle} \label{algo:line-spec}
\State $n^* \leftarrow (c \text{\bf { or} } \text{rand}() \geq \epsilon)$ $?$ $n^*:n^*+
(h^*(x_t) \neq  y_t)$ \label{algo:line-eps-sample}
\State $ n_c \leftarrow n_c + c$ \Comment{Increment if
  $\hat{h}_D$ did not use oracle} 
\State{Estimate acc. $a_{\hat{h}_D}$ of $\hat{h}_D$; use $p^*$
  derived from $S_j$ (\autoref{eqn:acc})} 
\If{ $a_{\hat{h}_D} < a^* + \tau_a$ {\bf or}\ $\frac{n^*}{n_c\cdot\epsilon} > \tau_{FP}$}  \label{algo:line-spec-exit}
\Statex {\Comment{Exit specialized classification}}
\State $j \leftarrow j + 1$  \Comment{Potentially start new epoch $j$}
\State go to Line \ref{algo:line-sample} \Comment{Go back to
  check if distribution has changed}
\Else
\State $S \leftarrow S \oplus [y_t]$; go to Line \ref{algo:line-spec} 
\EndIf
\end{algorithmic}
\caption{Windowed $\epsilon$-Greedy (WEG)}
\label{alg:opt-alg}
\end{algorithm}

The algorithm runs in three phases.
\begin{enumerate}
\item{\bf Window Initialization} [lines \ref{algo:line-sample} - \ref{algo:line-end-initialization}]
  identifies the dominant classes of the current epoch. To do so, we run
  the oracle on a fixed number $w_{min}$ (= 30 in our implementation)  of
  examples. The \textsc{DomClasses} helper identifies the dominant
  classes in the window as those that appear at least twice in the window. If
  the dominant classes are each within $\tau_r$ (= 2) of those
  of the previous epoch, we conclude the previous epoch is
  continuing and fold information collected on it into that for
  the current epoch $S_j$.
  
\item{\bf Template Selection} [lines \ref{algo:line-normal} - \ref{algo:template-selection-end}]
  Given a candidate set of dominant classes D, we estimate
  (\autoref{eqn:acc} below details precisely how) the
  accuracy of the cascaded classifier $\hat{h}_{D}$ for various template
  classifiers $h_i$ when specialized to $D$ and their current empirical
  probability skew $p^*$ derived from measured data $S_j$. Estimating
  these costs instead of explicitly 
  training the corresponding specialized classifiers $h_D$ is
  significantly cheaper. If the estimate is within a
  threshold $\tau_a$ (= 0.05 for object and scene recognition, and -0.05 for face
  recognition since the accuracy of oracle is higher) of
  the oracle, we produce the specialized model and go to the
  specialized classification phase. If not, we continue running the
  oracle on inputs and collecting more information on the incoming
  class skew.

\item{\bf Specialized Classification} [lines \ref{algo:line-spec-init} - \ref{algo:line-spec}]
  The specialized classification phase simply applies the current
  cascaded model $\hat{h}_D$ to inputs (Line \ref{algo:line-spec}) until
  it determines that the 
  distribution it was trained on (as represented by $D$) does not
  adequately match the actual current distribution. This determination is
  non-trivial because in the specialization phase, we wish to avoid
  consulting the oracle in order to reduce costs. However, the
  oracle is (assumed to be) the only unbiased source of samples from the
  actual current distribution.\\
  We therefore run the oracle in addition to the cascaded model with
  probability $\epsilon$ (= 0.01), as per the standard
  $\epsilon$-greedy policy for multi-arm bandits. Given the resulting
  empirical estimate $p^*$ of skew, we can again estimate the
  accuracy of the current cascade $\hat{h}_D$ as per
  \autoref{eqn:acc}. If the estimated accuracy of the cascade is too
  low, or if the classification results of $\hat{h}_D$ cascade are
  different from the oracle too often (we use a threshold $\tau_{FP}$
  = 0.5), we assume that the underlying distribution may have shifted
  and return to the Window Initialization phase.

\end{enumerate}

Finally, we focus on estimating the expected accuracy of the cascaded
classifier given the current skew $p$ of its inputs (i.e., the
fraction of its inputs that belong to the dominant class set). The accuracy of
cascaded classifier $\hat{h}_D$ can be estimated by:
\begin{align}
  \tiny
  a_{\hat{h}_D} & = p\cdot a_{in}+p\cdot e_{in\rightarrow out}\cdot a^* + (1-p)\cdot a_{out}\cdot a^*
\label{eqn:acc}
\end{align}
where $a_{in}$ is the accuracy of specialized classifier $h_D$ on $n$ dominant
classes, $e_{in\rightarrow out}$ is the fraction of dominant inputs that
$h_D$ classifies as non-dominant ones, and $a_{out}$ is the fraction of
non-dominant inputs that $h_D$ classifies as non-dominant (note that these
inputs will be cascaded to the oracle). We have observed previously
(\autoref{s:cnn}) that
these parameters $a_{in}$, $e_{in\rightarrow out}$, $a_{out}$ of specialized
classifier $h_D$ are mainly affected only by the size of the dominant
class $D$, not the identity of elements in it. Thus, we pre-compute these
parameters for a fixed set of values of $n$ (averaging over 10 samples
of $D$ for each $n$), and use linear interpolation for other $n$s at test time. 


%% file: cvpr/eval.tex
\section{Evaluation}
\label{s:eval}
\vspace{-6pt}
\begin{table*}[t]
\centering
\setlength{\tabcolsep}{3pt}
\scriptsize
\input{cvpr/table_synth}
\caption{Average accuracy and GPU latency of recognition over segments. For the
  segment column, each parenthesis indicates a segment of 5 minutes with the
  number of dominant classes and the skew.}
\label{tab-synth}
\end{table*}

\begin{table*}[t]
\scriptsize
\setlength\tabcolsep{3 pt}
  \centering
  \input{cvpr/table_video}
  \caption{Accuracy and average processing latency per frame on videos with
    oracle vs. WEG  (latencies are shown in ms). For additional insight, the last 5
    columns show key statistics from WEG usage.
  }
  \label{tab:video}
  \vspace{-12pt}
\end{table*}

\noindent
We implemented the WEG algorithm with a classification
runtime based on Caffe~\cite{jia2014caffe}. The system can be fed with videos to
produce classification results by recognizing frames. Our goal was to
measure both how well the large specialized model speedups of
\autoref{tab:cnns} translated to speedups in diverse settings and on long,
real videos. Further we wished to characterize the extent to which
elements of our design contributed to these speedups.

\subsection{Synthetic experiments}

\noindent
First, we evaluate our system with synthetically generate data in order to
study diverse settings. For this experiment, we generate a time-series of
images picked from standard large validation sets of CNNs we use.  Each test set
comprises of one or two segments where a segment is defined by the number of
dominant classes, the skew, and the duration in minutes. For each segment, we
assume that images appear at a fixed interval (1/6 seconds) and that each image
is picked from the testing set based on the skew of the segment. For an example
of a segment with 5 dominant classes and 90\% skew, we pre-select 5 classes as
dominant classes and pick an image with 90\% probability from the dominant
classes and an image with 10\% probability from the other classes at each time
of image arrival over 5 minutes duration. Images in a class are picked in a
uniform random way. We also generate traces with two consecutive segments with
different configurations to study the effect of moving from one context to the
other.

\autoref{tab-synth} shows the average top-1 accuracies and per-image processing
latencies using GPU for the recognition tasks with and without the specializer
enabled. The results are averaged over 5 iterations for each experiment. The
specializer was configured to use the compact classifiers O2 for objects, S2 for
scenes, and F2 for face recognition from~\autoref{tab:cnns}.

The following points are worth noting. (i) (Row 1 and it's sub-rows)
{\em WEG is able to detect and
exploit skews over 5-minute intervals and get significant speedups
over the oracle while preserving accuracy.} For the single segment
cases, the GPU latency speedup
per-image was 1.3$\times$ to 2.0$\times$, 1.3$\times$ to 2.4$\times$,
and 2.6$\times$ to 4.3$\times$, for object, scene, and face,
respectively. However, due to WEG's overhead these numbers are noticeably lower than
the raw speedups of specialized models (\autoref{tab:cnns}). When the
number of dominant classes increase, the specializer latency increases because
it alternates between exploration and exploitation to recognizes more dominant
classes. The latency also increases when the skew of dominant classes
decreases because specializer cascades more times to oracle model when using the
cascaded classifier. (ii) (Row 2)  {\em WEG is quite stable in handling random
inputs}, essentially resorting to the oracle so that accuracy and
latency are unchanged. (iii) (Rows 3 and 4) {\em WEG is able to detect
abrupt input distribution changes} as the accuracy remains comparable to oracle
accuracy, but with significant speedups when the distribution is
skewed (Row 4).

To further understand the limit of the WEG algorithm, we studied how frequently class
distributions can change before our technique stops showing benefit. We evaluated
face recognition with synthetic traces, changing the distribution every
10/20/30 sec. The WEG algorithm then yields speedups of 0.95/1.19/1.48$\times$
with roughly unchanged accuracy. For face recognition therefore, WEG
stops gaining benefit for distributions that lasts less than 20 secs.

\subsection{Video experiments}

\noindent
We now turn to evaluating WEG on real videos. However, we were unable to find a
suitable existing dataset to show off specialization.
We need several minutes or more of video that
contains small subsets of (the oracle model's) classes that may change
over time. In videos of real-world activity, this happens naturally; in popular benchmarks, not so much.
For example, the videos in YouTube Faces \cite{wolf2011face} are
short (average 6 sec, max 200 sec) and typically only contain one
person. Similarly, clips in UCF-101 \cite{soomro2012ucf101} have
mean length of 7.2 sec (max 71 sec) and focus on classifying
{\em actions} for which no oracle model exists. Finally, the
sports 1M dataset \cite{KarpathyCVPR14} assigns labels {\em per video}
instead of {\em per frame}.

As a consequence, we hand-labeled video clips from three movies, one TV show, and an interview and
manually labeled the faces in the videos
\footnote{The dataset is released at \url{http://syslab.cs.washington.edu/research/neurosys}.}.
The names of video clips with lengths
are listed in \autoref{tab:video}. Note that we used the entire videos for
Friends and Ellen Show, while we used a video chunk for the movies. For these
experiments, we used F2 as the compact classifier.

\autoref{tab:video} shows the average accuracies and average latencies
for processing a frame for 5 videos. We generated these by first extracting
all faces from the videos to disk using the Viola Jones detector. We
then ran WEG on these faces and measured the total execution
time. Dividing by the number of faces gave the average numbers shown
here. The most important point is that {\em even on real-world videos, WEG is able to
achieve very significant speedups over the oracle}, ranging from
2.6$\times$-11.2$\times$ (CPU) and  2.4$\times$-7.8$\times$ (GPU).

To understand the speedup, we summarize the statistics of WEG execution
in \autoref{tab:video}. ``Special rate'' indicates the percentage of time that specializer
exploits cascaded classifier to reduce the latency, while ``cascade rate''
reveals the percentage of time that a cascaded classifier cascades to the oracle
classifier, thus hurting performance. Higher special rate and lower
cascade rate yield more speedup. The cascade rate of ``Ocean's Eleven'' is
significantly higher than that of other videos. We investigated this and
found that the specialized compact CNN repeatedly made mistakes on one person in the
video, which led to a high cascade rate. ``Trans. special'' counts the
number of times WEG needed to switch between specialized and unspecialized
classification to handle the distribution changes and insufficient
exploration. The average dominant classes sizes (``dom. size'') show that the
real videos are skewed to fewer dominant classes than the
configurations used in the synthetic experiments. This explains why our
system achieved higher speedup on real videos than on synthetic
data. Overall, the statistics show that {\em the dataset exercise WEG
  features such as skew estimation, cascading and specialization}.

To understand better the utility of WEG's features, we performed an
ablation study: (a) We disable the
adaptive window exploration (Line 5-7 in \autoref{alg:opt-alg}), and use a fixed
window size of 30 and 60. (b) We use the skew of dominant classes in the input
distribution as the training skew for specializing compact CNNs instead of using
the fixed (50\%) training skew suggested in \autoref{s:cnn}. (c) We apply a
simple (but natural) criterion to exit from the specialized
classification phase: WEG now exits when the current skew is lower than the
skew when it entered into specialized classification phase instead of using the
estimated accuracy as soft indicator.

\begin{figure}[t]
  \centering
  \includegraphics[width=\linewidth]{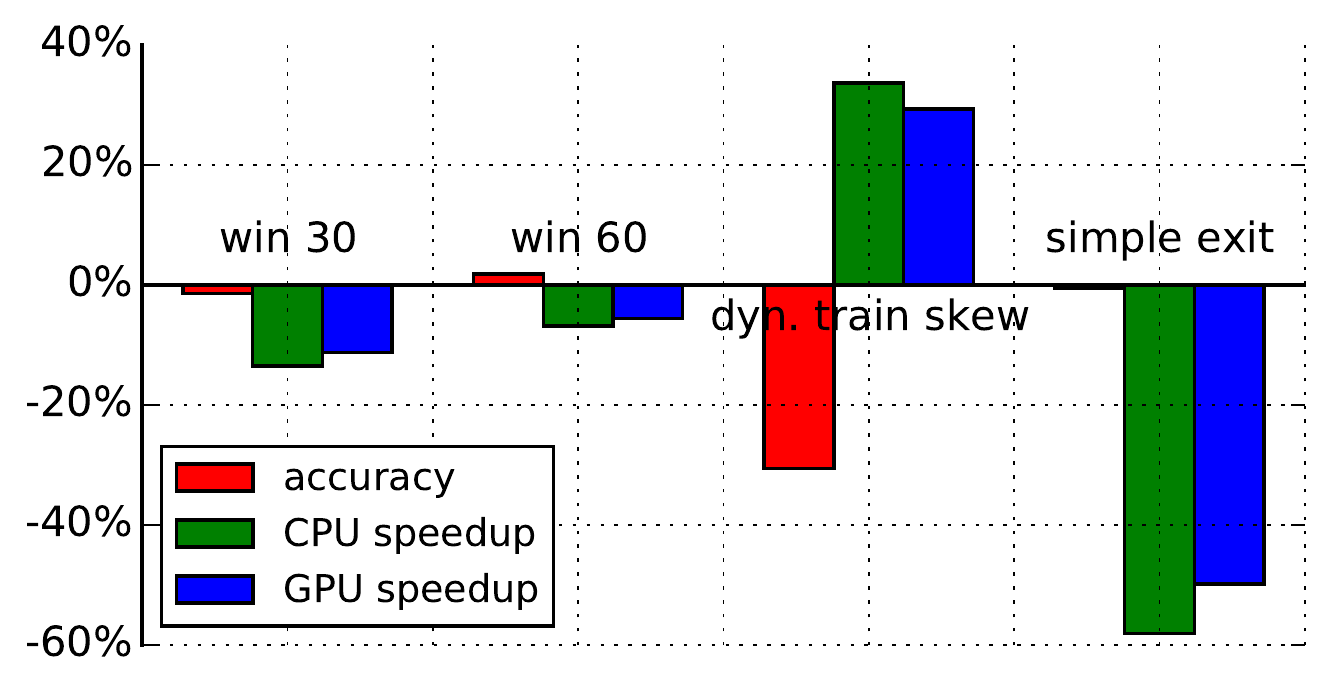}
  \caption{Change in accuracy (absolute difference) and speedup (relative) when individual features are disabled.}
  \label{fig:ablation}
  \vspace{-12pt}
\end{figure}

\autoref{fig:ablation} shows the comparison between these variants and WEG
algorithm in accuracy and CPU / GPU speedups when recognizing faces on Friends
video. In the figure we show the absolute differences in accuracy and relative
differences in CPU / GPU speedup. (a) Fixed window size (30 and 60) variants achieve
similar accuracy but lower speedup. As table \ref{tab:video} (``window
size'' column) shows, the adaptively estimated size for the window is
between 30 and 60. In general, too small a window fails to capture the
full dominant classes, yielding specializers that exit prematurely. Too
large a window requires more work by the oracle to fill up the
window. (b) Using variable rather than
fixed skew for training achieves more speedup, but suffers from 30\%
loss in accuracy. This is because
the training skew is usually very high. As discussed in
\autoref{s:cnn}, training on highly skewed data produces models
vulnerable to false positives in ``other'' classes. (c) The simple
exit variant achieves
almost comparable accuracy while the latency is more than 50\% higher than our
system. It demonstrates the value of our accuracy estimate in modeling the
accuracy of cascaded classifiers and to prevent premature exit from
the specialized classification phase. {\em In summary, the key design
  elements of WEG each have a role in producing fast and accurate results}.


%% file: cvpr/table_synth.tex
\begin{tabular}{cccccccccccccc}
\toprule
& \multicolumn{4}{c}{Object} & \multicolumn{4}{c}{Scene} & &\multicolumn{4}{c}{Face}\\
Segments & \multicolumn{2}{c}{disabled} & \multicolumn{2}{c}{enabled} & \multicolumn{2}{c}{disabled} & \multicolumn{2}{c}{enabled} & Segments & \multicolumn{2}{c}{disabled} & \multicolumn{2}{c}{enabled} \\
& acc(\%) & lat(ms)  & acc(\%) & lat(ms)  & acc(\%) & lat(ms)  & acc(\%) & lat(ms)  && acc(\%) & lat(ms)  & acc(\%) & lat(ms )  \\
\midrule
(n=5,p=.8) & 69.5 & 11.6 & 77.0 & 6.0 & 57.6 & 28.9 & 65.2 & 12.0 & (n=5,p=.8) & 95.2 & 28.7 & 95.1 & 9.2 \\
(n=10,p=.8) & 66.7 & 11.7 & 72.5 & 7.4 & 57.2 & 28.9 & 57.8 & 18.6 & (n=5,p=.9) & 97.0 & 28.6 & 96.2 & 6.7 \\
(n=10,p=.9) & 71.8 & 11.6 & 78.0 & 5.9 & 59.1 & 28.8 & 63.5 & 15.4 & (n=10,p=.9) & 95.4 & 28.5 & 94.3 & 11.0 \\
(n=15,p=.9) & 68.7 & 11.6 & 68.9 & 9.1 & 57.8 & 28.8 & 57.2 & 22.6 && &&& \\
\midrule
(random) & 68.1 & 12.1 & 68.1 & 11.5 & 59.1 & 28.9 & 59.1 & 28.8 & (random) & 95.9 & 28.5 & 95.9 & 28.5 \\
\midrule
(n=10,p=.9) & \multirow{2}{*}{67.9} & \multirow{2}{*}{11.8} & \multirow{2}{*}{70.2} & \multirow{2}{*}{9.1} & \multirow{2}{*}{57.0} & \multirow{2}{*}{28.8} & \multirow{2}{*}{56.0} & \multirow{2}{*}{22.6} & (n=5,p=.9) & \multirow{2}{*}{96.2} & \multirow{2}{*}{28.5} & \multirow{2}{*}{96.2} & \multirow{2}{*}{17.6} \\
+(random)
&&&&&&&&&+(random)&&&&\\
\midrule
(n=15,p=.9) & \multirow{2}{*}{70.6} & \multirow{2}{*}{11.6} & \multirow{2}{*}{73.9} & \multirow{2}{*}{7.8} & \multirow{2}{*}{61.1} & \multirow{2}{*}{28.7} & \multirow{2}{*}{63.0} & \multirow{2}{*}{17.1} & (n=10,p=.9) & \multirow{2}{*}{95.8} & \multirow{2}{*}{28.5} & \multirow{2}{*}{95.2} & \multirow{2}{*}{10.4} \\
+(n=5,p=.8)
&&&&&&&&&+(n=10,p=.8)&&&&\\
\bottomrule
\end{tabular}

%% file: cvpr/table_video.tex
\begin{tabular}{cc|ccc|ccc|ccccc}
  \toprule
  \multirow{2}*{video} & length & \multicolumn{3}{c|}{oracle} & \multicolumn{3}{c|}{WEG} & special & cascade & trans. & dom. & window \\
  & (min) & acc(\%) & CPU lat & GPU lat & acc(\%) & CPU lat & GPU lat & rate(\%) & rate(\%) & special & size & size \\
  \midrule
  Friends & 24 & 93.2 & 2576 & 28.97 & 93.5 & 538(\bf{$\times$4.8}) & 7.0(\bf{$\times$4.1}) & 88.0 & 7.5 & 51 & 2.8 & 41.8 \\
  Good Will Hunting & 14 & 97.6 & 2576 & 28.84 & 95.1 & 231(\bf{$\times$11.2}) & 3.7(\bf{$\times$7.8}) & 95.9 & 3.4 & 4 & 3.5 & 37.5 \\
  Ellen Show & 11 & 98.6 & 2576 & 29.26 & 94.6 & 325(\bf{$\times$7.9}) & 4.7(\bf{$\times$6.2}) & 93.7 & 4.8 & 19 & 1.7 & 47.4 \\
  The Departed & 9 & 93.9 & 2576 & 29.18 & 93.5 & 508(\bf{$\times$5.1}) & 6.9(\bf{$\times$4.2}) & 92.0 & 10.3 & 9 & 2.4 & 40.0 \\
  Ocean's Eleven / Twelve & 6 & 97.9 & 2576 & 28.97 & 96.0 & 1009(\bf{$\times$2.6}) & 12.3(\bf{$\times$2.4}) & 80.1 & 18.0 & 23 & 2.0 & 52.2 \\
  \bottomrule
\end{tabular}


%% file: cvpr/conclusion.tex
\section{Conclusion}
\label{s:conclusion}
\noindent
We characterize class skew in day-to-day video and show that the
distribution of classes is often strongly skewed toward a small number
of classes that may vary over the life of the video. We further show
that skewed distributions are well classified by much simpler (and
faster) convolutional neural networks than the large ``oracle'' models necessary
for classifying uniform distributions over many classes. This suggests
the possibility of detecting skews at runtime and exploiting them
using dynamically trained models. We formulate
this sequential model selection problem as the Oracle Bandit
Problem and provide a heuristic exploration/exploitation based
algorithm, Windowed 
$\epsilon$-Greedy (WEG). Our solution speeds up face recognition on TV episodes
and movies by 2.4-7.8$\times$ on a GPU (2.6-11.2$\times$ on a CPU) with little
loss in accuracy relative to a modern convolutional neural network.


%% file: cvpr/appendix.tex
\begin{table*}[!t]
\small
\centering
\input{cvpr/table_compact_cnn}
\caption{Architecture of compact models O1, O2, S1, S2, F1, and F2. The table specifies 
convolution layers as [kernel size, number of feature maps, stride, padding]; 
max-pooling layers as [kernel size, stride]; 
fully-connected layers as [output size].}
  \label{tbl:compact-models}
\end{table*}
\section{Generating Compact Models}
\autoref{tbl:compact-models} shows the six compact models used in the paper.
We created these models by systematically applying the following operations to
publicly available model architecture, such as AlexNet \cite{kriz} and VGGNet \cite{vggnet}:
(a) reduce the number of feature maps, or increase stride size in a convolution layer;
(b) reduce the size of a fully-connected layer; (c) merge two convolution layers
or a convolution layer and a max-pooling layer into a single layer.
The models are unremarkable except that they have fewer operations and layers than
original versions and hence run faster, yet achieve high
accuracy {\em when applied to small subsets of the original model's domain}.
In fact, how these architectures are derived is orthogonal to the WEG algorithm.
We have tested two fully automatic approximation techniques, tensor factorization
\cite{kim2016iclr} and representation quantization \cite{quantizedNN16} to
generate compact models as well.

\section{Dominant classes}

\begin{table}[!t]
\small
\centering
\setlength{\tabcolsep}{5pt}
\begin{tabular}{ccccccccc}
  \toprule
  index & $N$ & $a^*$ & $n$ & $p$ & $w$ & $k$ & $p_{in}$ & $p_{out}$\\\midrule
  1 & 1000 & 0.68 & 5 & 0.9 & 30 & 2 & 0.896 & 6.52E-5\\
  2 & 1000 & 0.68 & 10 & 0.9 & 30 & 2 & 0.558 & 6.53E-5\\
  3 & 1000 & 0.68 & 10 & 0.9 & 60 & 2 & 0.891 & 2.64E-4\\
  4 & 1000 & 0.68 & 10 & 0.7 & 60 & 2 & 0.789 & 4.80E-4\\
  5 & 1000 & 0.68 & 10 & 0.7 & 90 & 2 & 0.933 & 1.08E-3\\
  6 & 205 & 0.58 & 10 & 0.9 & 90 & 2 & 0.959 & 0.019\\
  7 & 205 & 0.58 & 10 & 0.9 & 90 & 3 & 0.872 & 1.31E-3\\
  8 & 205 & 0.58 & 0 & N/A & 90 & 3 & N/A & 9.29E-3\\\bottomrule
\end{tabular}
\vspace{5pt}
\caption{Probability $p_{in}$ and $p_{out}$ under various $N,a^*,n,p$ and $w,k$.
  In row 8, $n=0$ and $p=$ N/A indicates that the distribution has no
  skew and is uniform across all $N$ classes.}
\label{tab:window-support}
\end{table}

In the WEG algorithm, an important component is to decide the dominant classes
from a sliding window (function \textsc{DomClasses} in Algorithm 1). The
decision used in the algorithm is fairly simple: return the classes as dominant
classes that appear at least $k$ times in a sliding window $w$ of classification
result history from the oracle $h^*$, where $k$ is the minimum support number.

Here we use a simple model to analyze how to choose the minimum support number
$k$ for different window sizes $w$ in the WEG algorithm. Suppose $N$ is the
number of total classes classified by the oracle $h^*$ and the accuracy of $h^*$
is $a^*$. If the classification result from the oracle is wrong, the probability
that the oracle classifies the input to each of the other classes is assumed to
be equivalent. Suppose the input sequences are drawn independently from a skewed
distribution $T$ which has $n$ dominant classes with skew $p$. Denote the set of
dominant classes as $\mathcal{D}$ and set of non-dominant set as $\mathcal{O}$.
Based on these definitions, we can compute the probability of a single class
that the oracle classifier $h^*$ outputs given the input sequences.
Consider one dominant class $\ell \in \mathcal{D}$, the probability that it is
output by the oracle is:

\begin{align}
  \textrm{prob}(\ell \in \mathcal{D}) & = \frac{1}{n} \cdot \hat{p} 
  = \frac{1}{n} \left(p\cdot a^* + p(1-a^*)\frac{n-1}{N-1} \right. \nonumber \\
  & \left. + (1-p)(1-a^*)\frac{n}{N-1}\right)
  \label{eq:in-context-prob}
\end{align}
And the probability of a non-dominant class $\ell' \in \mathcal{O}$ that is
output by the oracle is:
\begin{align}
  \textrm{prob}(\ell' \in \mathcal{O}) &= \frac{1}{N-n} \left((1-p)a^* + (1-p)(1-a^*) \cdot
  \right. \nonumber \\
  & \left. \frac{N-n-1}{N-1}  + p(1-a^*)\frac{N-n}{N-1}\right)
  \label{eq:out-context-prob}
\end{align}

From \autoref{eq:in-context-prob} and \ref{eq:out-context-prob}, we can then
derive the probability of a dominant class or a non-dominant class that is
classified at least $k$ times in the window $w$ by using binomial
distribution. Intuitively, we hope the probability of a dominant class that
appear at least $k$ times ($p_{in}$) in the window be as high as possible, while
the probability of any non-dominant class ($p_{out}$) be as low as possible. We
considered different combinations of $N, a^*, n, p$ under different $w$ and $k$
settings, and computed $p_{in}$ and $p_{out}$. \autoref{tab:window-support}
shows how $p_{in}$ and $p_{out}$ changes under different settings. From the
table, we can tell that with an increase in the number of dominant classes and a
decrease in skew, we need to increase the size of window to have a higher
probability that the dominant classes can be detected in the window. However,
when the window size is larger, we also need to increase the minimum support
number $k$ (Row 6 and 7) to limit the probability that a non-dominant class
appears $k$ times. In addition, when there is no skew in the distribution (Row
8), the minimum support number $k$ is also effective at filtering out most of
the non-dominant classes. In practice, we set $k=2$ for $w < 90$ and $k=3$ for
$w\geq 90$.

%% file: cvpr/table_compact_cnn.tex
  \begin{tabular}{cc|cc|cc}
\toprule
    O1 & O2 & S1 & S2 & F1 & F2\\
\midrule
    \multicolumn{2}{c|}{input $3 \times 227 \times 227$} & \multicolumn{2}{c|}{input $3 \times 227 \times 227$} & \multicolumn{2}{c}{input $3 \times 152 \times 152$}\\\hline
    conv[11, 96, 4, 0] & conv[11, 64, 4, 0] & conv[11, 96, 4, 0] & conv[11, 64, 4, 0] & \multicolumn{2}{c}{conv[3, 64, 2, 1]}\\\hline
    \multicolumn{2}{c|}{pool[3, 2]} & \multicolumn{2}{c|}{pool[3, 2]} & \multicolumn{2}{c}{pool[2, 2]}\\\hline
    \multicolumn{2}{c|}{conv[5, 256, 2, 2]} & \multicolumn{2}{c|}{conv[5, 256, 2, 2]} & conv[3, 128, 1, 1] & conv[3, 128, 2, 1]\\\hline
    \multicolumn{2}{c|}{pool[3, 2]} & \multicolumn{2}{c|}{pool[3, 2]} & pool[2, 2] &\\\hline
    conv[3, 384, 1, 1] & conv[3, 256, 1, 1] & conv[3, 384, 1, 1] & conv[3, 256, 1, 1] & conv[3, 256, 1, 1] & conv[3, 128, 2, 1]\\\hline
    conv[3, 384, 1, 1] & conv[3, 256, 1, 1] & conv[3, 384, 1, 1] & conv[3, 256, 1, 1] & pool[2, 2] &\\\hline
    conv[3, 256, 1, 1] & & conv[3, 256, 1, 1] & & conv[3, 256, 1, 1] & conv[3, 256, 2, 1]\\\hline
    \multicolumn{2}{c|}{pool[3, 2]} & \multicolumn{2}{c|}{pool[3, 2]} & pool[2, 2] &\\\hline
    fc[4096] & fc[1024] & fc[2048] & fc[1024] & fc[2048] & fc[1024]\\\hline
    fc[4096] & fc[2048] & fc[2048] & fc[2048] & fc[2048] & fc[1024]\\\hline
    \multicolumn{2}{c|}{fc[1000]} & \multicolumn{2}{c|}{fc[205]} & \multicolumn{2}{c}{fc[2622]}\\
\bottomrule
\end{tabular}